\documentclass[11pt,a4paper]{article}
\usepackage[hyperref]{emnlp2018}
\usepackage{times}
\usepackage{footnote}
\makesavenoteenv{tabular}
\makesavenoteenv{table}
\usepackage{latexsym}
\usepackage{array,multirow,graphicx}
\usepackage{enumitem}
\usepackage{url}

\aclfinalcopy 

\title{Classifying Referential and Non-referential \textit{It} Using Gaze}

 \author{Victoria Yaneva, Le An Ha, Richard Evans, and Ruslan Mitkov\\ 
     Research Institute in Information and Language Processing,\\ University of Wolverhampton, UK\\        
{\tt\{v.yaneva, ha.l.a, r.j.evans, r.mitkov\}@wlv.ac.uk
}
}

\date{}

\begin{document}

\maketitle 
\begin{abstract}
When processing a text, humans and machines must disambiguate between different uses of the pronoun \textit{it}, including non-referential, nominal anaphoric or clause anaphoric ones. In this paper, we use eye-tracking data to learn how humans perform this disambiguation. We use this knowledge to improve the automatic classification of \textit{it}. We show that by using gaze data and a POS-tagger we are able to significantly outperform a common baseline and classify between three categories of \textit{it} with an accuracy comparable to that of linguistic-based approaches.  In addition, the discriminatory power of specific gaze features informs the way humans process the pronoun, which, to the best of our knowledge, has not been explored using data from a natural reading task. 

\end{abstract}

\section{Introduction}

Anaphora resolution is both one of the most important and one of the least developed tasks in Natural Language Processing \cite{lee2016qa}. A particularly difficult case for anaphora resolution systems is the pronoun \textit{it}, as it may refer to a specific noun phrase or an entire clause, or it may even refer to nothing at all, as in sentences 1 - 3 below\footnote{Extracted from the GECO corpus \cite{cop2016presenting}}.

\begin{enumerate} 
\setlength{\itemsep}{2pt}
  \setlength{\parskip}{2pt}
  \setlength{\parsep}{2pt}
\item \textit{``I couldn’t say exactly, sir, but \textit{it} wasn’t tea-time by a long way.''} (Pleonastic \textit{it} (non-referential)). 
\item \textit{“Now, as to this quarrel. When was the first time you heard of it?”} (Nominal anaphoric)
\item  \textit{“You have been with your mistress many years, is it not so?”} (Clause anaphoric\footnote{Some authors also distinguish other, less-common types of the pronoun it such as proaction, cataphoric, discourse topic, and idiomatic, among others \cite{evans2001applying}.}.)
\end{enumerate}

This phenomenon is not specific only to English; pronouns that can be used both referentially and non-referentially exist in a variety of language groups such as the pronoun \textit{`het'} in Dutch, \textit{`det'} in Danish, 
\textit{`ello'} in Spanish, \textit{`il'} in French, etc. 

In NLP, there has been active research in the area of automatic classification of \textit{it} 
during the past four decades but the issue is far from being solved (Section \ref{sec:relatedwork}). According to corpus statistics, the pronoun it is by far the most frequently used pronoun in the English language \cite{li2009identification}, and as recently as 2016, incorrect classification of  different cases of \textit{it} and their antecedents is highlighted as one of the major reasons for the failure of question-answering systems \cite{lee2016qa}. In the state of the art, the best approaches to classifying \textit{it} achieve between 2\% and 15\% improvement over a majority baseline when assigning examples of the pronoun to more than two classes. 

In contrast with the extensive research in NLP, very little is known about the way humans approach the disambiguation of \textit{it}. To the best of our knowledge, \newcite{foraker2007role}  is the only study on this subject that uses online measures of reading. It proves empirically that the pronoun \textit{it} is resolved more slowly and less accurately than gendered pronouns due to its ambiguity. So far there have been no studies investigating the subject using natural reading data as opposed to artificially created controlled sentences.

In this paper we approach these two problems as one and hypothesise that obtaining information on the way humans disambiguate the pronoun can improve its automatic classification. 

We propose a new method for classifying the pronoun \textit{it} that does not rely on linguistic processing. Instead, the model leverages knowledge about the way in which humans disambiguate the pronoun based on eye tracking data. We show that by using gaze data and a POS tagger we are able to achieve classification accuracy of three types of \textit{it} that is comparable to the performance of linguistic-based approaches and that outperforms a common baseline with a statistically significant difference. In addition, examining the discriminatory power of specific gaze features provides valuable information about the human processing of the pronoun. 
We make our data, code and annotation available at \url{https://github.com/victoria-ianeva/It-Classification}. The GECO eye-tracking corpus 
is available at \url{http://expsy.ugent.be/downloads/geco/}.

\section {Related Work}
\label{sec:relatedwork}

\paragraph{Gaze data in NLP} While eye-movement data has been traditionally used to gain understanding of the cognitive processing of text, it was recently applied to a number of technical tasks such as part-of-speech tagging \cite{barrett2016weakly}, detection of multi-word expressions \cite{rohanian2017using,yaneva2017cognitive}, sentence compression \cite{klerke2016improving}, complex-word identification  \cite{vstajner2017effects}, and sentiment analysis \cite{rotsztejn2018learning}. Eye movements were also shown to carry valuable information about the reader and were used to detect specific conditions affecting reading such as autism \cite{yaneva2018detecting,yaneva2016corpus} and dyslexia \cite{rello2015detecting}. 
The motivation behind these approaches is two-fold. First, eye tracking is already making its way into everyday use with interfaces and devices that feature eye-tracking navigation (e.g. Windows Eye Control\footnote{https://support.microsoft.com/en-gb/help/4043921/windows-10-get-started-eye-control}). Second, linguistic annotation by gaze is faster than traditional annotation techniques, does not require trained annotators, and provides a language-independent approach that can be applied to under-resourced languages \cite{barrett2016weakly}. This is particularly interesting for the case of non-referential pronouns, as the phenomenon exists in different languages. 
 
\paragraph{Classification of \textit{it}} The majority of the machine learning approaches to classifying the pronoun \textit{it} in different languages are based on linguistic features capturing token, syntactic and semantic context. Different papers report varying majority baseline metrics (between 50\% and 75\%) depending on the annotated corpora, and an improvement over the majority baselines of between 2\% and 15\% for classification of more than two classes of \textit{it} \cite{loaiciga2017disambiguating,uryupina2016detecting,lee2016qa,muller2006automatic,boyd2005identifying,hoste2007disambiguation,evans2001applying}. For example, \newcite{loaiciga2017disambiguating} train a bidirectional recurrent neural network (RNN) to classify three classes of \textit{it} in the ParCor corpus \cite{guillou2014parcor} and compare its performance to a feature-based maximum entropy classifier. The RNN achieves accuracy of 62\% compared to a majority baseline of 54\% and is significantly outperformed by the linguistic-feature classifier which obtained 68.7\% accuracy. \newcite{lee2016qa} compare several statistical models and report 75\% accuracy for four-class classification over a majority baseline of around 62\% using linguistic features and a stochastic adaptive gradient algorithm. They also report that experimenting with word embeddings did not lead to more accurate classification. So far, approaches using linguistic features still represent the state of the art in the classification of \textit{it}. 

\section{Data}
\label{sec:data}

\textbf{Corpus:} The eye-tracking data was extracted from the GECO corpus (see \newcite{cop2016presenting} for full corpus specifications) which is the largest and most recent eye-tracking corpus for English at present. The text of the corpus is a novel by Agatha Christie entitled ``The Mysterious Affair at Styles'', the English version of which contains 54,364 tokens and 5,012 unique types. 
The entire novel was read by 14 native English undergraduates from the University of Southampton using an eye-tracker with a sampling rate of 1 kHz. 


\textbf{Annotation scheme:} A total of 1,052 instances of \textit{it} were found in the corpus\footnote{This number does not include the possessive pronoun \textit{its}. There are also several tokenisation errors in the GECO corpus (e.g. ``it?..ah," and ``it...the" misidentified as single tokens.). These cases were excluded.}. Each of the instances of \textit{it} was annotated by two annotators and assigned to one of three categories: \textit{Pleonastic, Nominal anaphoric} and \textit{Clause anaphoric}, following the scheme used by \newcite{lee2016qa}. The annotators were free to view as much of the previous text as necessary to decide on a label. The inter-annotator agreement for the three categories was $\kappa$ = 0.636, p $<$ 0.0005, indicating substantial agreement between the annotators. This number corresponds to a percentage agreement of 77.47\% for the three categories and is comparable to the 81\% agreement reported in \newcite{lee2016qa}. We perceived adjudication between cases of disagreement (237 instances) to be extremely arbitrary, so those cases were excluded rather than resolved. Examples of such arbitrary cases include:

\begin{itemize}
\setlength{\itemsep}{2pt}
  \setlength{\parskip}{2pt}
\item \textit{"Sit down here on the grass, do. \underline{It}'s ever so much nicer."} (nominal vs. clause anaphoric)
\item \textit{“\underline{It}'s a jolly good life taking \underline{it} all round...if \underline{it} weren't for that fellow Alfred Inglethorp!"} (pleonastic vs. clause anaphoric)
\end{itemize}

The distribution of each class of the retained data by annotator is presented in Table \ref{tab:distribution}. We make the full annotations of both annotators available. 

\begin{table}
\small
   \begin{tabular}{|l|l|l|l|l|}\hline
   &Annotat. 1 & Annotat. 2& Final\\\hline
   Pleonastic& 339 (33\%)& 406 (38\%)&272 (33\%) \\\hline
   Nom. anaph.& 492 (46\%)& 527 (50\%)& 453 (56\%)\\\hline
   Clause anaph.& 221 (21\%)& 119 (11\%)& 89 (11\%)\\\hline
\end{tabular} 
\caption{Annotation categories}
 \label{tab:distribution}
\end{table}

\begin{table}
\small
\centering
\begin{tabular}{|l|l|l|l|l|}\hline
&Prev.  & ``It'' & Next & It + Next     \\\hline
Early  & 61.1  & 60.3 & 61.7      & 61   \\\hline
Medium & 60.9  & 60.5 & 60.6      & 61.2 \\\hline
Late   & 58.9  & 60.2 & 61        & 61.4\\\hline
\end{tabular}
\caption{Weighted F1 scores for an ablation study for different gaze feature groups over the Previous and Next word baseline (60.4)}
\label{tab:ablation}
\end{table}

\begin{table} [!htb]
\footnotesize
\begin{tabular}{|ll|l|l|l|}\hline
& & \rotatebox{90}{Ling}& \rotatebox{90}{Gaze} & \rotatebox{90}{Comb} \\\hline
\parbox{3mm}{\multirow{5}{*}{\rotatebox[origin=c]{90}{EARLY}}}  
& First\_Run\_Fixation\_Count  &&*&    \\
       & First\_Run\_Fixation\_\%      &&& *  \\
       & First\_Fixation\_Duration      &&&  \\
       & First\_Fixation\_Visited\_Count &&& \dag *\\
       & First\_Fix\_Progressive       &&&  \dag *\\\hline
\parbox{3mm}{\multirow{6}{*}{\rotatebox[origin=c]{90}{MEDIUM}}}  & Second\_Run\_Fixation\_Count    &&& \\
       & Second\_Run\_Fixation\_\%     &&&  * \\
       & Second\_Fixation\_Duration    &&*&  \dag *\\
       & Second\_Fixation\_Run         &&*&  * \\
       & Gaze\_Duration                &&& \dag \\\hline
\parbox{3mm}{\multirow{19}{*}{\rotatebox[origin=c]{90}{LATE}}} & Third\_Run\_Fixation\_Count   &&&   \\
       & Third\_Run\_Fixation\_\%      &&&  \dag \\
       & Third\_Fixation\_Duration     &&&  \dag * \\
       & Third\_Fixation\_Run         &&*&    \\
       & Last\_Fixation\_Duration     &&*&  \dag * \\
       & Last\_Fixation\_Run          &&&   \\
       & Go\_Past\_Time               &&&  \dag \\
       & Selective\_Go\_Past\_Time     &&&  \dag \\
       & Fixation\_Count              &&& \dag   \\
       & Fixation\_\%                &&*&  \dag   \\
       & Total\_Reading\_Time         &&&    \\
       & Total\_Reading\_Time\_\%    &&*& *    \\
       & Trial\_Fixation\_Count      &&&  *   \\
       & Trial\_Total\_Reading\_Time &&&     \\
       & Spillover              &&&          \\
       & Skip                    &&*&        \\\hline
   \parbox{3mm}{\multirow{20}{*}{\rotatebox[origin=c]{90}{LINGUISTIC}}} &    Word\_position  &+&&+ \\
& \# Preceding\_NPs\_in\_sentence       &&&  \\
& \# Preceding\_NPs\_in\_paragraph   &&&     \\
&\# Following\_NPs\_in\_sentence     &&&    \\
& \# NPs\_in\_the\_sentence               &&&     \\
& \# NPs\_in\_the\_paragraph              &&&  +   \\
& \# Following\_adject.\_in\_sentence &&&  \\
&Previous\_verb                         &+&&               \\
&Following\_adjective                   &+&& +              \\
&Following \_verb                       &+&& +              \\
&POS in posit. L4, L3, L2, L1    &+&+\footnotemark& +           \\
&POS in posit.
R4, R3, R2, R1            &+&+& +  \\
& \# Following\_complementisers            &+&& +   \\
&An\_adjective\_before\_the\_next\_NP      &+&& + \\
&Words\_until\_next\_complementiser         &+&& +     \\
&Words\_until\_next\_infinitive              &+&& +    \\
&Words\_until\_next\_preposition            &&&  +  \\
&Words\_until\_next\_ing\_verb              &+&& +    \\
&A\_compl.\_before\_the\_next\_NP    &&& \\
&Immediately\_preceding\_preposit. &&&   \\\hline
 \parbox{3mm}{\multirow{4}{*}{\rotatebox[origin=c]{90}{BASIC}}} 
 &Previous\_word                      &+&+&+                  \\
&Next\_word &+&+&+ \\
&Word\_length &&+&+\\
&Punctuation &&& +\\\hline
\end{tabular}
\footnotetext{Except L4 and L3}
\caption{List of features and their inclusion in the different models. + refers to linguistic data, * to added values for the \emph{It + Next} region, and \dag  to gaze features for the previous word region. The features that do not have marks in the last three columns were not retained in any of the three best models.}
\label{tab:features}
\end{table}

\section{Experiments}

\paragraph{Overview} In order to test the extent to which gaze data can help the classification of different cases of \textit{it}, we trained and compared three separate classifiers. The first classifier is based on gaze features, the second one is based on linguistic features and finally, we trained a combined classifier using both gaze and linguistic features. We compared the performance of these classifiers to a majority baseline of 55.7 and to another baseline obtained by using the tokens surrounding the pronoun (previous and next word) as features (60.4). We also experimented with adding word embeddings\footnote{300-dimensional vectors from Google News obtained through word2vec: https://code.google.com/archive/p/word2vec/} for the surrounding tokens as features. While the full exploration of word embeddings for the classification of \textit{it} remains outside of the scope of this work, it would be interesting to explore whether the embeddings add value to the models by encoding information that was not otherwise captured. 

\paragraph{Gaze features}  We use the gaze features as provided in GECO and we average the data from all 14 readers per token. We extract gaze data for each case of \textit{it}, as well as for the preceding and following word. The full list of gaze features used in the experiments can be seen in Table \ref{tab:features}. 

Different eye-tracking measures (usually divided into \textit{early} and \textit{late}) are indicative of different aspects of cognitive processing. Early gaze measures such as \textit{First\_Fixation\_Duration} give information about the early stages of lexical access and syntactic processing, while late measures such as \textit{Total\_Reading\_Time} or \textit{Number\_of\_Fixations} give information about processes such as textual integration and disambiguation (see \newcite{rayner2012psychology} for a review). In Table \ref{tab:features}, the distinction between \textit{Early, Medium} and \textit{Late} gaze features is mainly based on the run during which the fixations were made (i.e. whether the eyes were passing through the text for the first, second or third time). For each run we report count measures, percentage measures (as part of the trial) and duration measures (in milliseconds). Additional late features reported include \textit{Last\_Fixation\_Duration} and \textit{Last\_Fixation\_Run} (the run during which the last fixation in a given region occurred),  \textit{Total\_Reading\_Time} (in msec and \%), and \textit{Trial\_Fixation\_Count} (the overall number of fixations within the trial). \textit{Go\_Past\_Time} refers to the summation of all fixation durations on the current word during the first pass. \textit{Spillover} refers to the duration of the first fixation made on the next word after leaving the current word in the first run. Finally, a word is considered skipped if no fixation occurred during the first run (\textit{Skip}). A complete legend explaining each feature can be found within the corpus metadata.

An ablation study on the contributions of individual groups of gaze features towards the classification of \textit{it} is presented in Table \ref{tab:ablation}. 

\paragraph{Linguistic features} We implemented a set of features originally proposed by \newcite{evans2001applying} and subsequently used extensively 
in the studies presented in Section \ref{sec:relatedwork}. In terms of features and categories of \textit{it}, the study by \newcite{evans2001applying} is the most fine-grained one we could find, classifying 7 categories of \textit{it} with 69\% accuracy. The set of features from \newcite{evans2001applying} is presented in Table \ref{tab:features}. These features synthesize information based on corpus studies of the pronoun \textit{it} and thus aim to capture positional, part-of-speech and proximity information, as well as specific patterns of usage. For example, \newcite{evans2001applying} notes that pleonastic pronouns rarely appear immediately
after a prepositional word and that complementisers or adjectives often follow pleonastic instances. Another pattern that distinguishes the pleonastic use of \textit{it} is associated with certain sequences of elements such as `adjective + noun phrase' and `complementiser + noun phrase' \cite{evans2001applying}. Therefore, the linguistic features proposed by \newcite{evans2001applying} and used in our experiments make possible the utilization of corpus-based knowledge for the automatic classification of \textit{it}.

\paragraph{Classification} We use simple logistic regression as implemented in WEKA \cite{hall2009weka} with 10-fold cross validation and a random seed parameter 20. Since logistic regression is an interpretable method, we are able to assess the performance of individual features and gain insight into the psycholinguistic processing of the pronoun.

We experimented with gaze features for the individual words but as gaze data is inherently very noisy, we found that smoothing the features by adding the ones that correspond to the pronoun and the next word stabilized the results. Adding the gaze features for the previous word significantly reduced the performance but using them separately in a model with the added \textit{It + Next} features maximised our results. For the role of individual features in the models see Table \ref{tab:features}.

\begin{table}
\small
\label{tab:results}
\begin{tabular}{llll}
&\textbf{P}                          & \textbf{R} & \textbf{F1}       \\
Baselines&&&\\\hline
Majority    baseline             &        &      &   55.7    \\
Previous + Next word          & 62.1   & 63.9 & 60.4      \\
\\
Embeddings&&&\\\hline
Prev. + Next Embed.& 63.1 & 64.4 & 62.1 \\
\\
Linguistic models&&&\\\hline
Full feature set            & 63.4   & 66   & 63.2        \\
Best linguistic  & \textbf{66.7}   & \textbf{68.8} & \textbf{66.1*}   \\
Best linguistic + Embed. & \textbf{66.9} & \textbf{68.8} & \textbf{67.2*}  \\
\\
Gaze-based models&&&\\\hline
Basic + POS & 63.3 & 64.5 & 62.2 \\
Select. Gaze + Basic & 65.8   & 66.8 & 64.2 \\
Select. Gaze + Basic + POS & \textbf{66.6}   & \textbf{67.9} & \textbf{65.6*} \\
S. Gaze + Bas. + POS + Embed. & \textbf{66.3} & \textbf{68.8} & \textbf{66.7*}  \\
\\
Combined model&&&\\\hline
Best Gaze + Ling & \textbf{71}   & \textbf{71.5} & \textbf{68.8*}   \\
Best Gaze + Ling. + Embed.& 67.5& 69.3 & 671* \\
\end{tabular}
\caption{Precision, Recall and Weighted F1 for the various classifiers. The * symbol marks statistical significance compared to the baseline model of Previous + Next Word (60.4).}
\end{table}

In order to account for class imbalance we compute and report a weighted F1 score, as opposed to the harmonic mean between precision and recall. First, the F1 for each class is weighted by multiplying it by the number of instances in the class. Then the F1 scores for all classes are summed up and divided by the total number of instances. The resulting weighted F1 score is lower than the traditionally reported mean F1 score, but it represents the effects of class imbalance more accurately.

\section{Results and Discussion}

The results from the classification experiments presented in Table \ref{tab:results} have implications for language processing by both humans and machines.

From the NLP perspective, the potential of gaze data to not only improve but also, to a certain extent, substitute text processing approaches is an exciting new frontier. Our results show that the gaze-based classifier performs on par with the one using linguistic features and both of them perform significantly better\footnote{Gaze + Basic + POS: \textit{p} = 0.029, 95\% CI (0.509; 9.858) ; Best Linguistic: \textit{p} = 0.0017, 95\% CI (1.01; 10.34); Best Gaze + Ling: \textit{p} = 0.0004, 95\% CI (3.75; 12.99). The CI indicate the difference in \%} than the baseline of 60.4. An improvement of 13\% over the majority baseline is achieved when combining the two but this difference is not a significant improvement over the individual best classifiers. A possible reason for this is that the gaze data and linguistic features encode similar information about the disambiguation of \textit{it} and adding them together leads to overlap instead of complementation. In all three classifiers, the clause-anaphoric class was consistently predicted with lowest accuracy (Table \ref{tab:confusion}), which is not surprising given that it only accounts for 11\% of the retained data. 
In line with the observation of \newcite{lee2016qa} (Section \ref{sec:relatedwork}), the embeddings do not show a stable contribution. In our case, this is likely related to the small amount of data, to which we add 300 dimensions per word. 

Overall, the improvement achieved by the classifiers is comparable  with the current state-of-the-art (Section \ref{sec:relatedwork}). It is important to note that this is the first study to use text from the domain of literature and that this may have influenced the extraction of the linguistic features. At the same time, literature can be regarded as a more challenging domain than the declarative texts used in previous research, owing to the creative use of language.

\begin{table}
\small
\label{tab:confusion}
\begin{tabular}{|l|l|l|l|} \hline
NomAnaph & ClauseAnaph & Pleon & \\\hline
395 &2 & 56 & NomAnaph \\\hline
53 & 11 & 25 & ClauseAnaph \\\hline
93  & 3 & 176 & Pleon\\\hline
\end{tabular}
\caption{Confusion matrix for the best combined model (Weighted F1 = 68.8)}
\end{table}

From a psycholinguistic perspective, we provide evidence that, indeed, the three classes of \textit{it} are processed differently. We observe that medium and late gaze features related to disambiguation are more discriminative than the early ones. For example, measures such as first fixation duration were not included in any of the models, while revisits as late as the third run (the third time the eyes pass over the region of interest) occurred in these regions and provided a strong signal. Particularly useful features were the durations of the second and last fixations, as well as the information about the run (pass) during which they occur.

The significance of these features in the best classifiers somewhat contradicts the ablation study presented in Table \ref{tab:ablation}. According to that table, early processing features for the preceding and next words are expected to outperform the late ones. A possible explanation for this are predictability and spillover effects, as the pronoun \textit{it} is both highly predictable and easy to skip, because of its high frequency and shortness. Indeed, the gaze features from the it-region itself are not as useful as the ones from the surrounding words.

The results from this study showed that: i) gaze features encode information about the way humans disambiguate the pronoun \textit{it}, ii) that this information partially overlaps with the information carried by linguistic features, and that iii) gaze can be used for automatic classification of the pronoun with accuracy comparable to that of linguistic-based approaches. In our future work we will attempt to identify specific patterns of cognitive processing for the individual classes, as well as explore factors related to the readers.

\section{Conclusion}

We presented the first study on the use of gaze data for disambiguating categories of \textit{it}, exploring a wide range of gaze and linguistic features. The model based on gaze features and part-of-speech information achieved accuracy similar to that of the linguistic-based model and state-of-the-art systems, without the need for text processing. Late gaze features emerged as the most discriminative ones, with disambiguation effort indicators as late as third pass revisits.

\bibliography{emnlp2018}
\bibliographystyle{acl_natbib_nourl}

\appendix

\end{document}